\documentclass{article}

\usepackage{arxiv}

\usepackage[utf8]{inputenc} % allow utf-8 input
\usepackage[T1]{fontenc}    % use 8-bit T1 fonts
\usepackage{hyperref}       % hyperlinks
\usepackage{url}            % simple URL typesetting
\usepackage{booktabs}       % professional-quality tables
\usepackage{amsfonts}       % blackboard math symbols
\usepackage{nicefrac}       % compact symbols for 1/2, etc.
\usepackage{microtype}      % microtypography
\usepackage{lipsum}		% Can be removed after putting your text content
\usepackage{graphicx}
\usepackage{doi}
\usepackage[square,numbers]{natbib} %for square references

\title{Computational behavior recognition in child and adolescent psychiatry : A statistical and machine learning analysis plan}

\makeatletter
\let\@fnsymbol\@arabic
\makeatother
\author{
Nicole N.~Lønfeldt
\thanks{Child and Adolescent Mental Health Center, Copenhagen University Hospital – Mental Health Services CPH, Hellerup, Denmark }\hspace{0.15cm}\textbf{\textsuperscript{*}}
\and 
\textbf{Flavia D.~Frumosu} 
\thanks{Department of Applied Mathematics and Computer Science, Technical University of Denmark, Kongens Lyngby, Denmark}\hspace{0.15cm}\textbf{\textsuperscript{*}}
\and 
A.-R.~Cecilie~Mora-Jensen
\footnotemark[1] 
\and 
Nicklas Leander~Lund
\footnotemark[2]
\and 
Sneha~Das
\footnotemark[2] 
\and 
A.~Katrine~Pagsberg
\footnotemark[1] \hspace{0.05cm}\textbf{\textsuperscript{,}}\hspace{-0.05cm}
\thanks{Department of Clinical Medicine, Faculty of Health
and Medical Sciences, University of Copenhagen, Copenhagen, Denmark}\hspace{0.15cm}\textbf{\textsuperscript{,}}\hspace{-0.05cm}
\thanks{Department of Clinical Biochemistry, Hospital Glostrup - University Hospital, Glostrup, Denmark \hfill \break \textbf{\textsuperscript{*} Equal contributor,} \textbf{Correspondence}: \texttt{nicole.nadine.loenfeldt@regionh.dk}}
\and 
Line~K.~H.~Clemmensen
\footnotemark[2] 
}

\renewcommand{\headeleft}{Lønfeldt \textit{et al.}}
\renewcommand{\shorttitle}{{}}

\hypersetup{
pdftitle={Computational behavior recognition in child and adolescent psychiatry: A statistical and machine learning analysis plan},
pdfsubject={cs.CV},
pdfauthor={Nicole N.~Lønfeldt, Flavia D.~Frumosu, A.-R.~Cecilie~Mora-Jensen, Nicklas Leander~Lund, Sneha~Das, A.~Katrine~Pagsberg, Line~K.~H.~Clemmensen},
pdfkeywords={machine learning,
visual signals from video data,
human behavioral coding,
children and adolescents,
obsessive-compulsive disorder},
}

\begin{document}
\maketitle

\begin{abstract}
\textbf{Motivation}\\
Behavioral observations are an important resource in the study and evaluation of psychological phenomena, but it is costly, time-consuming, and susceptible to bias. Thus, we aim to automate coding of human behavior for use in psychotherapy and research with the help of artificial intelligence (AI) tools. Here, we present an analysis plan.\\
\textbf{Methods}\\ %if any
Videos of a gold-standard semi-structured diagnostic interview of 25 youth with obsessive-compulsive disorder (OCD) and 12 youth without a psychiatric diagnosis (no-OCD) will be analyzed. Youth were between 8 and 17 years old. Features from the videos will be extracted and used to compute ratings of behavior, which will be compared to ratings of behavior produced by  mental health professionals trained to use a specific behavioral coding manual. We will test the effect of OCD diagnosis on the computationally-derived behavior ratings using multivariate analysis of variance (MANOVA). Using the generated features, a binary classification model will be built and used to classify OCD/no-OCD classes.\\
\textbf{Discussion}\\ %if any
Here, we present a pre-defined plan for how data will be pre-processed, analyzed and presented in the publication of results and their interpretation. A challenge for the proposed study is that the AI approach will attempt to derive behavioral ratings based solely on vision, whereas humans use visual, paralinguistic and linguistic cues to rate behavior. Another challenge will be using machine learning models for body and facial movement detection trained primarily on adults and not on children. If the AI tools show promising results, this pre-registered analysis plan may help reduce interpretation bias.
Trial registration: ClinicalTrials.gov - H-18010607 
\end{abstract}

% keywords can be removed

\section*{Introduction}
Observations of human behavior comprise an important clinical and research tool within psychology and psychiatry. For example, motor unrest and slowness are symptoms of psychiatric disorders \cite{american2013diagnostic} \cite{world2004icd}. Facial expressions, eye contact, proximity, touch, and gaze provide information about mental states such as emotions, attention, and cognitive effort, as well as interpersonal processes such as alliance, affection, and reciprocity \cite{benito2021therapist}\cite{feldman2012cib}\cite{hudson2001parent}\cite{mcleod2004therapy}. 
While behavioral coding bypasses the problem of self-report bias, disadvantages also exist. Human behavioral coding is a costly, time-consuming task, and it is susceptible to other types of bias. Thus, we aim to use AI tools that will reduce human labor as well as increase the speed and reliability of coding human behavior using landmark detection i.e., extracting features from videos. 

\section*{Methods}

\subsection*{Participants and setting}
A total of 37 participants between the ages of 8 and 17 years are included in this study. Participants include 25 youth with OCD and 12 youth without a psychiatric diagnosis from an larger case-control study and randomized clinical trial – TECTO \cite{pagsberg2022family}. All participants were screened for psychiatric disorders by mental health professionals (psychologists, medical doctors with psychiatric training and graduate-level psychology students). Diagnoses of patients were established by psychiatrists or specialized clinical psychologists.  

\subsection*{Measures}
The Kiddie Schedule for Affective Disorders and Schizophrenia (K-SADS) is a semi-structured interview used to screen for previous and current psychiatric symptoms and  diagnoses in children and adolescents between the ages of 6 and 18 \cite{puig1986kiddie}. Diagnostic categories have a subscale of items. Items are scored on a 3-point scale (1 = symptom not present, 2 = subclinical, 3 = clinical). If an item is rated 3, guidelines instruct the interviewer to complete the relevant disorder supplement to establish whether the full criteria of a diagnosis has been met. Items can also be scored with zero to indicate that no information is available. Items are rated for parent and child responses. The total item score takes the parent and child response as well as the clinician’s expertise into account. 
Before inclusion in TECTO, all participants and their parents were interviewed with the K-SADS interview. Here, we used the clinician’s total score for current symptoms. The scores from the depression and mania chapters were used to establish whether the video footage is of participants discussing current psychiatric symptoms. Some participants scored 2 or 3 for the depression chapter, while all the participants scored 1 for the mania chapter.  

\subsection*{Video data source} 
\begin{figure}[h!]
 \centering
  \includegraphics[width=11cm]{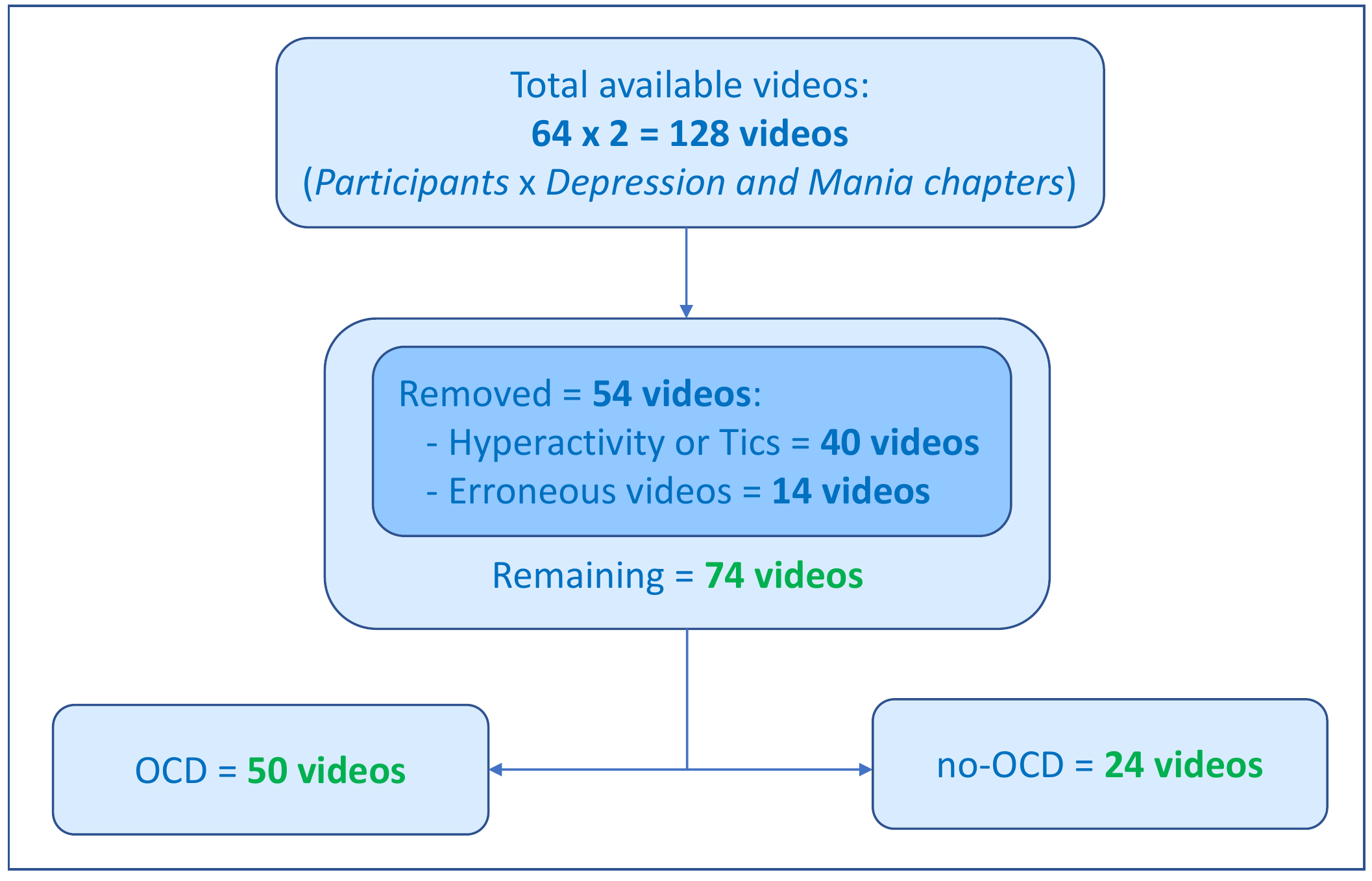}
  \caption{Video selection for the study}
  \label{fig:videos}
  \end{figure}

We will extract visual/motor features from videos of K-SADS interviews of children. Figure \ref{fig:videos} shows how videos were selected for this study. TECTO aims to include about 30\% fewer controls than patients and started including patients before controls. Therefore, there are more patient videos available than control videos. Videos containing a depression and mania chapter were available for 64 participants. The videos were initially recorded with the purpose of quality assurance and training of mental health professionals. As they were not filmed for video processing purposes, the data quality is not optimal. The videos are shot from different angles and contain a varying number of people in the frame. Many videos are missing for several reasons. First, some patients did not consent to being filmed before being included in TECTO. Second, many videos were discarded to follow data protection policies in the clinic before they could be moved to research folders. In total, 7 participants or 14 videos were removed due to missing data on clinical measures of interest, overexposure of the video or other technical problems. Videos of children who received a K-SADS score over one on the motor unrest item of the attention deficit hyperactivity chapter, the stereotyped/repetitive behavior item in the autism spectrum disorder chapter or the motor tic item were excluded to avoid videos that may include increased movements unrelated to outcomes of interest. In this initial work, we decided not to exclude children with vocal tics, attention deficits not problems with social reciprocity. These symptoms may be rated correctly by humans and the computational models. Finally, it could be argued that we were not conservative enough and that other symptoms should also be removed. For example, vocal tics may increase the vocalization scores, but vocalization should capture socially motivated speech and laughter. Other symptoms should be captured by behavioral codes. A lack of social reciprocity may result in decreased gaze, attention and positive affect, and attention deficits are expected to be reflected in lower attention codes. Nonetheless, this study is a first step and in future work we will test successful models in more symptom-diverse populations. The second author (CM) with expertise in psychopathology and conducting K-SADS interviews, identified the start and stop times of the depression and mania chapters of the K-SADS interview videos. 

\subsection*{Coding Interactive Behavior (CIB) adolescent version}
The CIB is a global rating system for coding interactions between two partners on a scale from one to five, in which half-points also can be assigned (e.g., 1.5 or 3.5) \cite{feld1998cib}. One indicates that the behavior is low in intensity, frequency, and duration or not present. Five indicates that the behavior is high in intensity, frequency, and duration. Separate codes are assigned to the parent, child and the dyad as a unit. While the original coding system was developed for parent-child interactions, it has also been used for interactions between children and adults other than the children's parents \cite{klein2007mothers}. The adolescent version was designed for youth between the ages of 12 and 18, but many of the codes appear in versions developed for infancy to adolescence for continuity \cite{feldman2012cib}. Several individual codes can be averaged to represent theoretically meaningful constructs such as social engagement \cite{feldman2012cib}. The manual recommends a minimum of three minutes of interaction for global scores and 10 second slices for micro-codes. In this work, we sought to decrease human labor and machine processing time to test our method and thus, applied global scores to 30-second clips. We were careful to choose codes which involve behaviors or mental processes that could be observed in 30 second intervals and not complex codes that require more context. This study will focus on child behavior and thus, will only use child codes. We chose codes that are relevant for evaluating child engagement and emotional states. These constructs are valuable in and of themselves, but we also believe they will contribute to assessing therapeutic alliance, parent-child synchrony, and child distress during exposure-based therapy in future work using other videos (Clinicaltrials.gov - H-18010607). We will use the following seven child codes, as they are relevant to interactions between children and mental health professionals:
\vspace{0,2cm}
\begin{enumerate}
  \item \textbf{Gaze} - child's eyes look in direction of the interviewer.
  \item \textbf{Attention} - child listens to the interviewer and responds to the interviewer in relevant ways.
  \item \textbf{Positive affect} - verbal and nonverbal expressions that indicate the child is feeling relaxed and comfortable, including smiling, laughter, warm tone of voice and relaxed body posture.
  \item \textbf{Negative emotionality} - verbal and nonverbal expressions of negative emotions (e.g. sadness and anger) including frowning, closed body posture (crossed arms), yelling, and crying. 
  \item \textbf{Vocalization} - speech and other vocal noise produced with the goal of social interaction. Frequent, consistent, and clear speech is scored high.
  \item \textbf{Anxiety} - overt signs of nervousness such as  frequent looks to the camera or interviewer, enthusiasm that is incongruent with the context, long silences, actions such as pulling hair, fidgeting, nail biting, emotional lability, or statements expressing anxiety, fear or nervousness.
  \item \textbf{Activity-level/arousal} - level of expressed energy in the way the child talks and acts. Calm speech and movements receive a low score, whereas passionate, frantic or excited speech and activity receives a high score. Energetic speech can be marked by rapidity, loudness and excess. 
\end{enumerate}
\vspace{0,2cm}
Gaze, attention, positive affect, and vocalization can be combined into a composite score representative of child social engagement \cite{feldman2012cib}.
\subsection*{Human behavioral coding procedure}
One of the first authors and the second author (NL and CM), a psychologist and medical doctor specializing in psychiatry, respectively, will code the videos. The first author was certified by the developer of the CIB and subsequently trained the second author. They reached a rater agreement of Kappa = 89\% for a separate set of videos coded for training. They will hold regular meetings to prevent coder drift. Coders will not code videos in which they are the interviewers. 
Blinding human coders to diagnostic status cannot be achieved as the discussion of symptoms is inherent to the interview. However, we attempt to blind coders to the extent possible. Thus, participants received a new ID number that does not reveal whether they are cases or controls and video clips received a code that does not reveal whether they contain the depression or mania chapter. CM has had contact with many of the participants in the TECTO study and is not blind to much of their clinical information, which may bias CM to score patients more negatively than controls or patients with higher clinical severity more negatively than patients with lower clinical severity. Thus, we estimated the sample size needed to test for a  difference in means between CM’s codes and NL’s codes at a significance level of 0.05 (two-tailed), a power of 80\%, a mean difference of 1 (the minimum allowed difference for agreement) and a standard deviation (SD) of 0.9 \cite{lebowitz2016salivary} using G*Power \cite{faul2007g}. Results indicate that we would need a sample of 14 per rater to detect a mean difference of one. If we assume a less conservative SD of 0.11 \cite{feldman2007mother} and a mean difference of one, a sample size of four would be needed to achieve a power of 80\%. Thus, 14 videos (33\%) will be coded by both human coders to test bias in the unblinded coder and inter-rater reliability. For the entire analysis, in total, 44 videos will be coded by each of the human coders.

\section*{Statistical Analysis Plan}
\subsection*{Processing pipeline}
\subsubsection*{Pre-processing video and annotations}
The two video clips from the depression and mania chapters from each participant varied in length. The shortest video was approximately 40 seconds in duration. We extracted 30 seconds from each video from the middle section and coded this. The middle section was deemed the most representative of the diagnostic chapter as we expect this to be where the child is more active, talkative and expressive. The diagnostic chapters start with the interviewer posing a question and by the end of the chapter the potential emotion-provoking stimuli (a question about a psychiatric symptom) may be distant or no longer novel and so children may be more relaxed. 

\subsubsection*{Video features}
To extract features from the videos, pre-trained machine learning models focusing on landmark detection and facial expression recognition are used. These models automatically track the position and movement of the head, hands, fingers, face, and torso. A computational facial expression recognition model is used to classify facial expressions from images of human faces into emotion categories: anger, disgust, fear, happiness, sadness, surprise and neutral. With such models, we can automatically track specific movements or positions as well as emotions that humans use to assign behavioral codes. Table \ref{table:models} presents an \textit{a priori} description of the computationally-tracked actions that we believe will contribute to specific CIB codes. 

\begin{table}[h!]
\caption{Models used for body movement tracking and facial expression recognition}
\label{table:models}
\begin{tabular}{lllll}
\hline
\textbf{CIB codes/constructs} & \textbf{Tracked actions} & \textbf{Class machine learning models} &  &  \\ \hline
1 Gaze & \begin{tabular}[c]{@{}l@{}} Head and eyes \\ positioning/angles \end{tabular} & Gaze tracking &  &  \\ \hline
2 Attention & \begin{tabular}[c]{@{}l@{}}Nodding head/turning head \\ from side to side (yes or no), \\ Limited movement (focus)\end{tabular} & \begin{tabular}[c]{@{}l@{}}Body landmark detection, \\ Face landmark detection\end{tabular} &  &  \\ \hline
3 Positive affect & Facial expression for happiness & Facial Expression Recognition &  &  \\ \hline
4 Negative emotionality & \begin{tabular}[c]{@{}l@{}} Facial expression \\ for sadness or anger \end{tabular} & Facial Expression Recognition &  &  \\ \hline
5 Vocalization & \begin{tabular}[c]{@{}l@{}}Openness of mouth,\\ Ratio between mouth corners\\ and lower, upper lips\end{tabular} & Face landmark detection &  &  \\ \hline
6 Anxiety & \begin{tabular}[c]{@{}l@{}}Activity frequency (high), \\ Facial expression for fear, \\ Facial expression for disgust\end{tabular} & \begin{tabular}[c]{@{}l@{}}Body landmark detection,\\ Face landmark detection\\ Facial Expression Recognition\end{tabular} &  &  \\ \hline
7 Activity & \begin{tabular}[c]{@{}l@{}}Hands: away from the torso,\\ touching head, touching arm,\\ crossing arms\\ Torso: frequent movements\\ Head: up, down, \\ left, right movements\end{tabular} & \begin{tabular}[c]{@{}l@{}}Body landmark detection, \\ Face landmark detection\end{tabular} &  &  \\ \hline
\end{tabular}
\end{table}

\begin{flushleft}
As the data were not acquired for the purpose of video analysis, the videos vary in terms of characteristics since different cameras were used to film the interviews. Thus, the computationally-derived ratings of behavior will initially be represented in percentages, e.g., how many times a certain event has been observed, divided by the number of frames from the video. Subsequently, the percentages are transformed into scores ranging from 1 to 5 to match the human-rated CIB scores.
\end{flushleft}

%Trick to make it nice
%\vspace{1cm}

\section*{Statistical analyses}
\subsection*{Statistical analysis I}
First, we will estimate the agreement between the computationally-derived scores and the corresponding human-derived, CIB scores using a Cohen's kappa, for easy comparison to human agreement, and a correlation plot for visualization. In other words, the reliability between the human coders and the features extracted by the machine learning models are checked. The seven scores can be treated as composite scores and reduced to a smaller number of scores: social engagement, negative emotionality, anxiety and activity.

\subsection*{Statistical analysis II}
Furthermore, we will test for significant differences at a level of $\alpha$ = 0.05 in the computationally-derived scores between the diagnostic groups (OCD and no-OCD). For this analysis, the composite scores are used in a MANOVA. The composite scores will also be used to test differences between the depression and mania chapters of the OCD and no-OCD groups. If necessary, a correction of the p-values will be performed using the Benjamini-Hochberg procedure controls \cite{benjamini_hochberg1995} with a false discovery rate (FDR) of 20\%.

\subsection*{Statistical analysis III}
Using the generated features data, a binary classification model can be trained in which the classes are OCD and no-OCD classes. Examples include tree based models (e.g., random forest, extreme gradient boosting) and logistic regression. These models can be used for predicting OCD and no-OCD classes on unseen video data. 
A tree approach is preferred in this context as it also produces feature importance as output, which can help to identify the best features for capturing the OCD/no-OCD classes. \\Regarding model validation strategies, depending on the selected model different statistics will be reported, e.g., inspection of qq-plots of the residuals, Out-of-Bag (OOB) error. The data is imbalanced as OCD patients outnumber no-OCD participants. Thus, we consider using either oversampling or undersampling strategies to deal with the class imbalance. As performance metrics we will report the F1-Score and area under the curve (AUC) computed from the receiver operator characteristic curve (ROC). \\
For estimating the generalization error of the models, we plan to use either leave one out cross validation (LOOCV) or 5-fold cross validation, as deemed necessary. If relevant, stratified cross validation will be used.

\section*{Discussion}
\subsection*{Strengths}
Here, we present a pre-defined plan for how data will be pre-processed, analyzed and presented in the publication of results and their interpretation. If the AI tools show promising results, this pre-registered analysis plan may help reduce interpretation bias. Automation of coding human behavior will save clinicians and researchers of human behavior time and resources. An additional advantage is that coding will become more consistent and less prone to human error and bias.
A challenge for the proposed study is that the AI approach will attempt to derive behavioral ratings based solely on vision, whereas humans use visual, paralinguistic and linguistic cues to rate behavior. If successful, these AI tools would transcend human abilities. In future work, visual AI tools can be supplemented with audio tools to improve automatic ratings. Another challenge will be using machine learning models for body and facial movement detection trained primarily on adults and not on children. Children and adults differ in the expression of mental states, i.e., attention and emotions, and thus, more work is needed on younger populations \cite{abbasi2022statistical}.

 \subsection*{Limitations}
In this initial study, we chose to remove subclinical symptoms of motor unrest, stereotyped/repetitive behaviors and motor tics and motor unrest. The reason behind this choice is to ease the comparison between the OCD/no-OCD groups, as we have few subjects, we want to minimize unnecessary noise.
This decision may decrease the influence of bias of the human coder with knowledge of the patients' diagnoses as the other human coder without this knowledge may code more similarly to the machine that also lacks this knowledge. In this case, agreement between the human coders would decease and we may falsely conclude that machine-human agreement is at least as high as human-human agreement. This decision also resulted in removing many children from the sample and will limit the generalizability of our findings. 

\section*{Conclusion}
The current predefined analysis plan will help to limit bias for reported results in a future publication. If the obtained results from this analysis plan are promising, we will be one step closer to automated human behavioural coding within the context of child psychiatry. 

\section*{Competing interests}
The authors declare that there are no competing interests for this work.

\section*{Acknowledgements}
The current work is funded by the Novo Nordisk Foundation (grant number: NNF19OC0056795). The authors would also like to thank research assistant Anders Buch Thuesen for his work while temporarily supporting the project.

\bibliographystyle{abbrvnat}
\bibliography{references}  %%% Uncomment this line and comment out the 

\end{document}